\newif\ifanonymous
\anonymousfalse

\newcommand{\phantomName}{%
    \ifanonymous
        \emph{MAFrameworkX}%
    \else
        \emph{Phantom}%
    \fi
}%

\documentclass[sigconf,nonacm]{aamas}

\usepackage{balance} 
\usepackage{pifont}
\usepackage{multirow}
\usepackage{amsmath}

\setcopyright{ifaamas}
\acmConference[AAMAS '23]{Proc.\@ of the 22nd International Conference
on Autonomous Agents and Multiagent Systems (AAMAS 2023)}{May 29 -- June 2, 2023}
{London, United Kingdom}{A.~Ricci, W.~Yeoh, N.~Agmon, B.~An (eds.)}
\copyrightyear{2023}
\acmYear{2023}
\acmDOI{}
\acmPrice{}
\acmISBN{}

\acmSubmissionID{395}

\title{\phantomName{} - A RL-driven multi-agent framework to model complex systems}

\author{Leo Ardon}
\affiliation{%
  \institution{J.P. Morgan AI Research}
  \country{}
}
\email{leo.ardon@jpmorgan.com}

\author{Jared Vann}
\affiliation{%
  \institution{J.P. Morgan AI Research}
  \country{}
}
\email{jared.vann@jpmorgan.com}

\author{Deepeka Garg}
\affiliation{%
  \institution{J.P. Morgan AI Research}
  \country{}
}
\email{deepeka.garg@jpmorgan.com}

\author{Thomas Spooner}
\affiliation{%
  \institution{Sutter Hill Ventures}
  \country{}
}
\email{spooner10000@gmail.com}

\author{Sumitra Ganesh}
\affiliation{%
  \institution{J.P. Morgan AI Research}
  \country{}
}
\email{sumitra.ganesh@jpmorgan.com}

\begin{abstract}
    Agent based modeling (ABM) is a computational approach to modeling complex systems by specifying the behavior of autonomous decision-making components or agents in the system and allowing the system dynamics to emerge from their interactions. Recent advances in the field of Multi-agent reinforcement learning (MARL) have made it feasible to study the equilibrium of complex environments where multiple agents learn simultaneously.
    However, most ABM frameworks are not RL-native, in that they do not offer concepts and interfaces that are compatible with the use of MARL to learn agent behaviors. In this paper, we introduce a new open-source framework, \phantomName, to bridge the gap between ABM and MARL. \phantomName{} is an RL-driven framework for agent-based modeling of complex multi-agent systems including, but not limited to economic systems and markets. The framework aims to provide the tools to simplify the ABM specification in a MARL-compatible way - including features to encode dynamic partial observability, agent utility functions, heterogeneity in agent preferences or types, and constraints on the order in which agents can act (e.g. Stackelberg games, or more complex turn-taking environments). In this paper, we present these features, their design rationale and present two new environments leveraging the framework.
\end{abstract}


\newcommand{\BibTeX}{\rm B\kern-.05em{\sc i\kern-.025em b}\kern-.08em\TeX}

\usepackage{caption}
\usepackage{subcaption}
\usepackage{float}
\graphicspath{ {./figures/} }

\begin{document}


\pagestyle{fancy}
\fancyhead{}


\maketitle 


\section{Introduction}

Agent based modeling (ABM) is a paradigm to model complex systems in a bottoms-up manner by specifying the behavior of autonomous decision-making components in the system (or agents); and allowing the system dynamics to emerge from their interactions. Drawing upon their real-world counterparts they seek to model, agents assess the state of the world and make decisions that will affect the rest of the system inducing the emergence of non-trivial phenomena. ABM offers several advantages over traditional differential equations modeling often used to study system dynamics. First, the description of problems is more natural because the real world is composed of autonomous entities. Second, it offers flexibility in the way the agents are modeled, with the option to replicate the heterogeneity of behaviors observed in real life. 

Recent advances in the field of Reinforcement Learning (RL) have brought another dimension to the study of complex multi-agent systems with the introduction of an autonomous learning component to the ABM paradigm. This line of research seeks to study the equilibrium of such non-stationary environments where multiple agents learn at the same time, by playing against or with each other. Multi-agents reinforcement learning (MARL) techniques have been applied to autonomous vehicles, cooperative agents systems and trading simulators \citep{canese2021multi}. 

However, most frameworks for agent-based modeling are not RL-native, in that they do not offer concepts and interfaces that are compatible with the use of MARL to learn agent behaviors in a specified ABM. Our goal with \phantomName{} is to bridge the gap between ABMs and MARL. \phantomName{} is an RL-driven framework for agent-based modeling of complex multi-agent systems such as economic systems and markets. It leverages the power of MARL to automatically learn agent behaviors or policies, and the equilibria of complex general-sum games. To enable this, the framework provides tools to specify the ABM in MARL-compatible terms - including features to encode dynamic partial observability, agent utility / reward functions, heterogeneity in agent preferences or types, and constraints on the order in which agents can act.


In this paper, we elaborate on the architecture and design of the \phantomName{} framework and provide details about the main features and their rationale\footnote{We provide the code in the supplementary materials and will open source the framework.}. Finally, we show how this framework can be used to model complex environments such as markets like the digital ads market or even operational problem in the supply chain environment. We also evaluate the scalability of the proposed framework by running experiments involving a high number of agents.


\section{Principal Features}

\subsection{Partial Observability}

The agents in an ABM interact by sharing information with each other, that can affect their behavior and eventually lead to uncovering interesting phenomena. However, in many real-world applications not all the information shared across the system is available for all the agents to consume e.g. a bidder entering an auction does not know how much its competitors are willing to bid, a market maker might only be able to observe the pricing inquiries it receives and its own transactions, a customer using a ride-sharing app might only see local drivers. 

Most real-world problems have a strong component of partial observability and it was therefore crucial for our framework to support partially observable environments seamlessly and with the guarantee that there will be no information leakage among the agents. We propose in our framework, a customizable network model to design complex relationships between the different agents in the system and we offer a safe mechanism to ensure that only the specified information is shared with the other agents, guaranteeing true partial observability.

\subsubsection{Network Model}\label{sec:network-model}
\hfill

In \phantomName, we model the relationship between agents in the system as a network or graph where each vertex / node represents an agent and each edge represents an open line of communication between two agents. One of our main desiderata for the framework was the ability to support complex and dynamic connectivity patterns between the agents. For this reason, we decided to treat the network component as a first-class citizen of the framework. The network can be seen as the physical layer on which the information is sent through, which means that two agents will only be able to communicate if an edge exists between the two vertices representing them. This property of the framework turns out to be particularly powerful to express partial observability. 

The network being a component on its own, it is possible to encapsulate logic to update the network dynamically and replicate as closely as possible real-world interactions. For example, in a global currency (FX) market, agents might enter and exit the market at different times depending on their time-zone. In a ride-sharing market, the connectivity of a customer to drivers depends on geographical proximity which might vary with time as the agent moves. These examples require dynamic or stochastic networks which can be implemented in \phantomName{} by extending a well-defined network interface. Users can implement their logic in a custom \textit{Network} class and update the network topology at any point during the experiments, with the guarantee that two agents will be able to share information if and only if there are connected.

\begin{figure}[t]
  \centering
  \centerline{\includegraphics[scale=0.6]{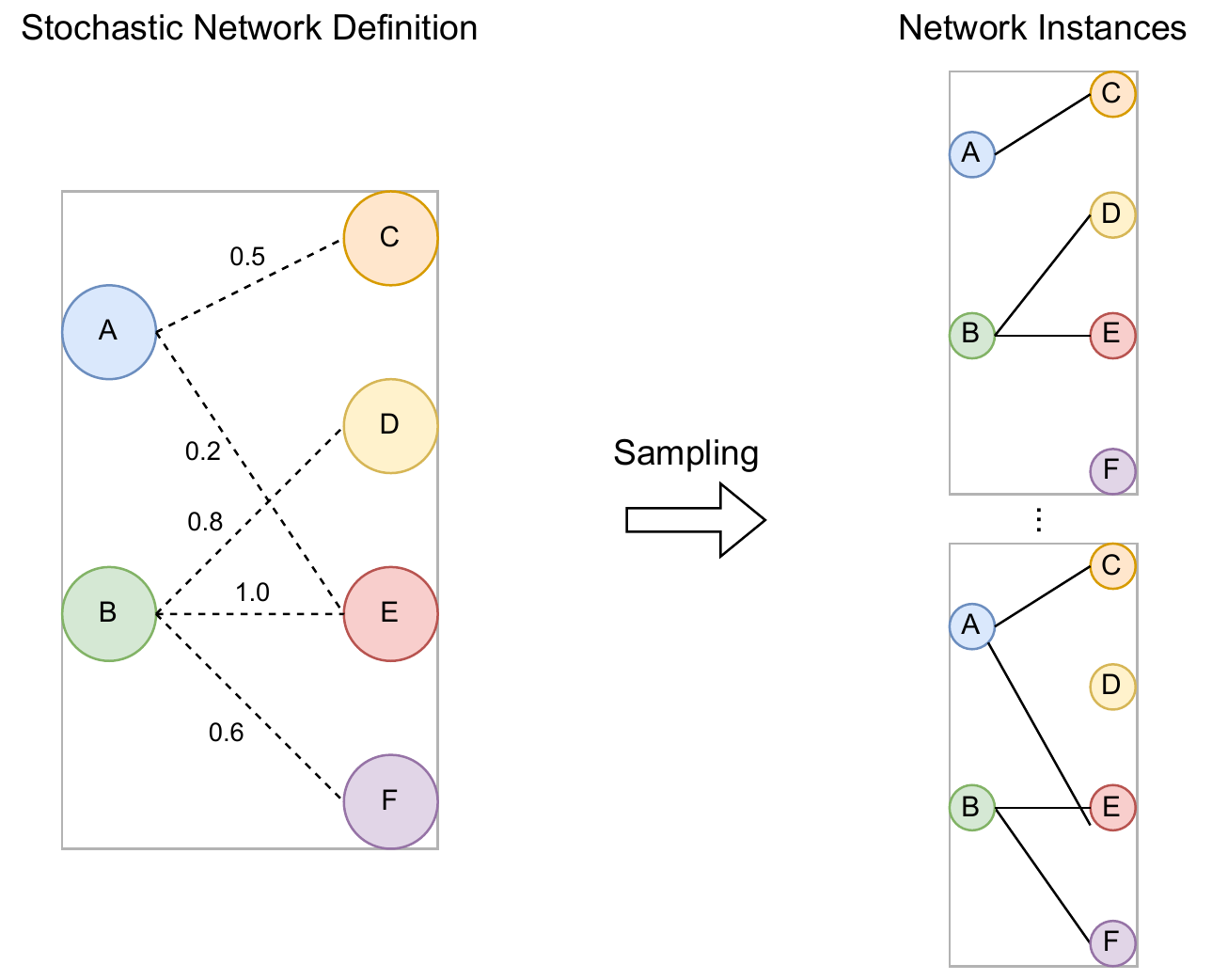}}
  \captionsetup{justification=centering}
  \caption{Overview of the Stochastic Network where each edge is associated with a probability of connecting two agents. The network can be sampled during the experiment and yields different structures of the relationships between the agents. This type of network can be used to model complex system with dynamic relationships.}
  \label{fig:stochastic-network}
  \Description{Sampling mechanism of the Stochastic Network component allowing the agents to be exposed to various connectivity configurations.}
\end{figure}

As part of the framework, we provide two different implementations of the network that already cover a range of use cases \citep{lowe2017multi, ganesh2019reinforcement, ardon2021towards}. The first one is a \textbf{static network} where the connectivity between the agents is defined upfront and remains static throughout training and simulation. The second implementation, more robust, is a \textbf{stochastic network} where each edge connecting two agents, is associated with a probability of existing. The network can be `re-sampled` during RL-training between the episodes, to yield a new structure which can impact the behaviors of the agents in the system (Figure~\ref{fig:stochastic-network}). Adding stochasticity in the connectivity among agents helps prevent the MARL algorithms from overfitting to a specific network topology and is particularly useful to generalize the learned policies over a range of possible connectivity patterns when the actual graph is not known \textit{a priori} \citep{ardon2021towards}.

\begin{figure*}[t]
    \centering
    \begin{subfigure}[b]{0.25\textwidth}
        \centering
        \includegraphics[width=\textwidth]{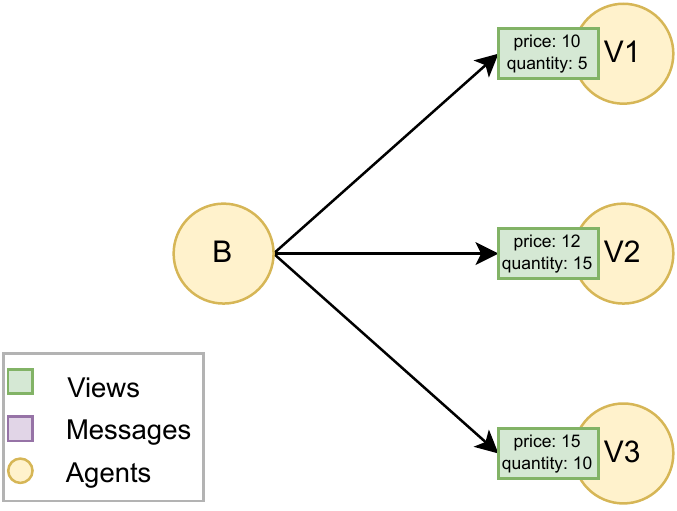}
        \captionsetup{justification=centering}
        \caption{\textbf{B} wants to buy $10$ units of products at the lowest price possible. It consumes the data exposed in the \textit{View} of each vendor}
        \label{fig:views1}
    \end{subfigure}
    \hfill
    \begin{subfigure}[b]{0.25\textwidth}
        \centering
        \includegraphics[width=\textwidth]{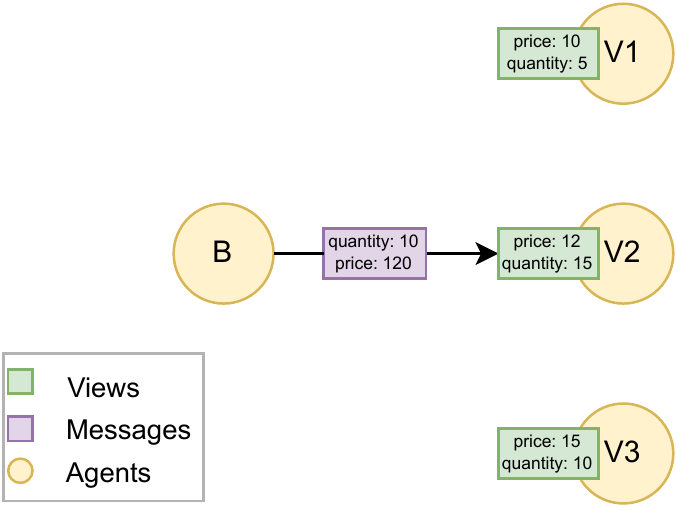}
        \captionsetup{justification=centering}
        \caption{\textbf{B} decides to buy from \textbf{V2} because it offers the cheapest price for the quantity required. \textbf{B} sends a \textit{Message} to \textbf{V2} to buy the product.}
        \label{fig:views2}
    \end{subfigure}
    \hfill
    \begin{subfigure}[b]{0.25\textwidth}
        \centering
        \includegraphics[width=\textwidth]{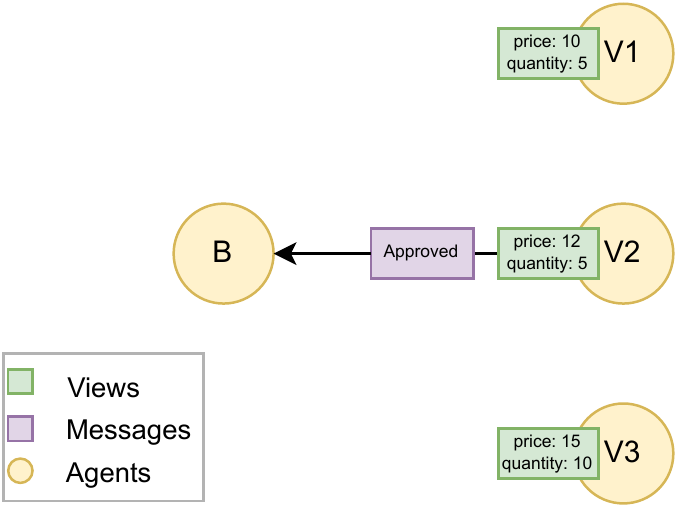}
        \captionsetup{justification=centering}
        \caption{\textbf{V2} replies to \textbf{B} by sending a \textit{Message} confirming the sell. \textbf{V2} also updates the \textit{View} exposed to \textbf{B} to reflect the new quantity of product available.}
        \label{fig:views3}
    \end{subfigure}
    \captionsetup{justification=centering}
    \caption{An interaction example between agents to illustrate the concepts of Message and View.}
    \label{fig:messages-views}
    \Description{Example of how Views and Messages can be used to pass information between agents.}
\end{figure*}

\subsubsection{Messages and Views}\label{sec:messages-views}
\hfill

In \phantomName, we offer two mechanisms to share information with a neighbour agent. The first one, which we qualify as `active`, take the form of a \textit{Message} intentionally sent at a time $t$ from one agent to another. This active way of sharing information ensures that the information has been consumed by the receiver of the message. A message is triggered by an event in the system, such as a new time step or the reception of another message. The emission of a new message is often associated with the decision making process choosing the information the agent wants to share.

On the other hand, the second mechanism to share information is referred to as `passive`. The agent simply exposes specific information for others to consume but does not actively send it. We use \textit{Views} to encapsulate the data to be shared. Each agent generates a customized view for each of its neighbors with only the information required. The views are regularly updated but no notification is sent to the other agents. It is entirely up to them to decide if and when they consume that information. The collection of views from all the neighbors of a given agent represents the context of that agent at a given time $t$ and can be used to make decisions. \textit{Views} are particularly useful when the data exposed changes frequently and does not necessarily require an action from the other agent; instead of sending a message for every change the \textit{View} will be updated without much processing from the system leading to higher overall performances.

As opposed to some other frameworks using message buses to expose an agent's state to the other participants in the system, \phantomName{} enforces the communication to only occur through the edges of the underlying network characterizing the connectivity between agents. In effect, \textit{Messages} and \textit{Views} can only be shared with an agent's neighbors when there exists an edge connecting the two nodes representing the agents. This aspect of the framework, as well as the ability to have different \textit{Views} for different neighbors, are designed to easily encode the partial observability associated with many real-world problems. Implementing such a property in a subscription based model over a message bus would require ad-hoc validation logic to evaluate whether an agent can subscribe to a particular topic. The direct use of the underlying network to pass \textit{Message} and to access \textit{Views} greatly simplifies the implementation of the multi-agent system for the end user of the framework.

To make the concepts of \textit{Messages} and \textit{Views} more concrete, we provide a simple example (Figure~\ref{fig:messages-views}). Let us consider a system where multiple vendors are trying to sell a product and a buyer wants to buy this product at the lowest possible price. Each vendor, due to their own internal finances, offers the product at a different price. The price and the quantity of product available is typically the kind of information that will be exposed in a \textit{View} by the vendor agent, for connected buyers to consume during their decision making process. To decide whom to trade with, the buyer accesses the view exposed by each of the vendors and evaluates the price and the quantity of product available. Based on this information, the buyer makes an educated decision on which vendor to buy the product from and sends him a \textit{Message} with the quantity of product required and the payment. The vendor replies by returning another message approving the transaction.

\subsection{Heterogeneity of Learned Agent Behaviors}

Specifying the behaviors of agents in the system, and how they evolve, is one of the crucial tasks in specifying an ABM for a domain and often requires hand-coding of known strategies in classical ABM approaches. While \phantomName{} supports taking actions from a hand-crafted (fixed or evolving) policy, it is also natively geared towards supporting MARL as an approach to train the policies of the agents. 

A MARL-driven approach to building an agent-based model requires specification of agents' rewards or utility functions; the agent behaviors emerge from the learning process as each agent tries to maximize its reward in the presence of other learning agents. In \phantomName, the \textit{Agent} definition includes specification of its observation and action spaces, as well as a reward function. 

\subsubsection{Types}
\hfill

The agent reward or utility function is typically parameterized by a vector of values referred to as the \textit{Type} - which in effect, implies that the agent class is associated with a space of possible reward functions rather than a single fixed one. Each Type value specifies a particular instance of the reward formulation and an associated learned behavior. For example, a market maker agent in a financial market might want to maximize profit and loss (PnL) while minimizing risk - this could be encoded as a parameterized reward function $PnL - \gamma*Risk$ where the Type parameter $\gamma$ encodes the agent's preferences regarding the trade-off (or its risk-aversion). 

There are two advantages to this approach: (1) A single agent class could be used to learn a range of behaviors depending on the instantiation of Type parameters, making the ABM specification compact. (2) It is also often the case, that while the modeler might have sufficient domain knowledge about the general form of the reward function, it might not have direct knowledge of the exact form for each agent. \phantomName{} allows the modeler to specify the general, parameterized form of the reward function - treating the Types as hyper-parameters that could be specified or calibrated.


\subsubsection{Supertypes}
\hfill

\begin{figure*}[h]
\begin{minipage}{\textwidth}
  \centering
  \begin{subfigure}[t]{0.35\textwidth}\centering
    \raisebox{-0.5\height}{\includegraphics[height=3.25in]{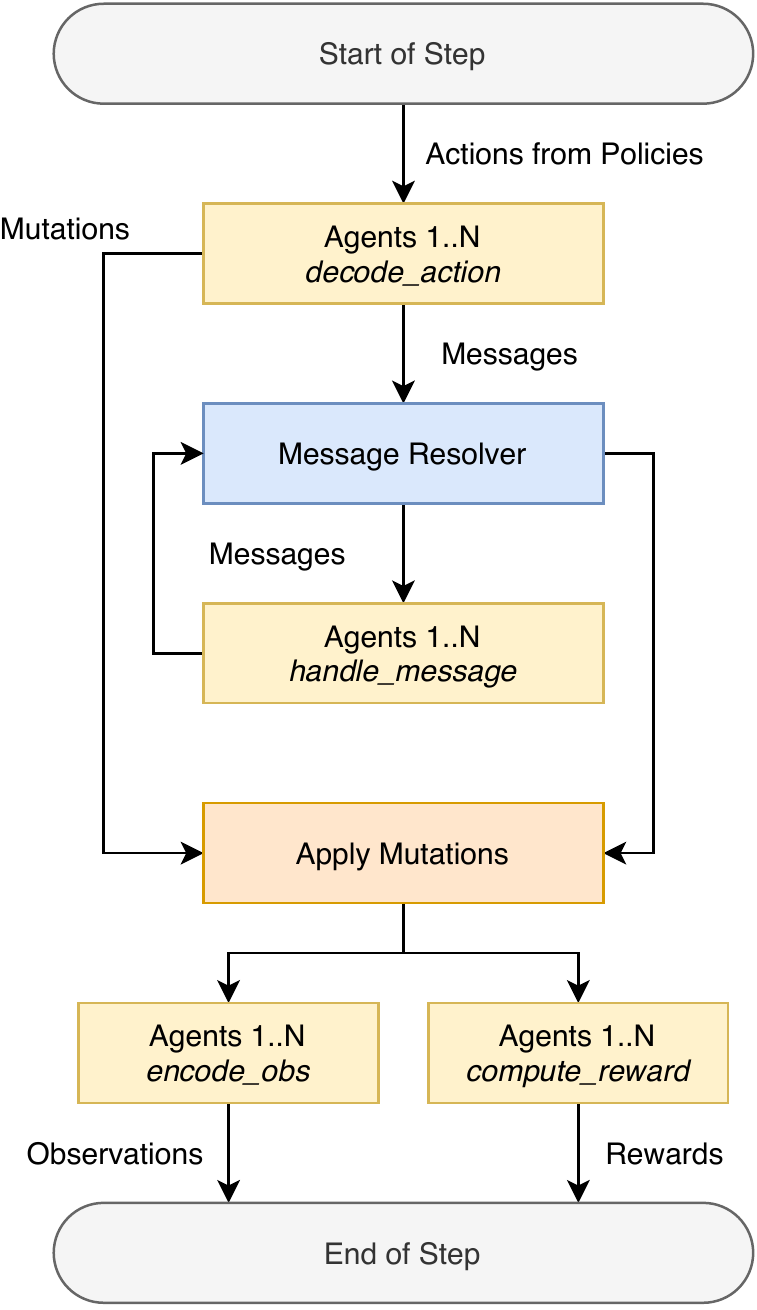}}
    \caption{Step Breakdown}
    \label{fig:step_breakdown}
  \end{subfigure}
  \begin{subfigure}[t]{0.54\textwidth}\centering
    \raisebox{-0.5\height}{\includegraphics[height=2.35in]{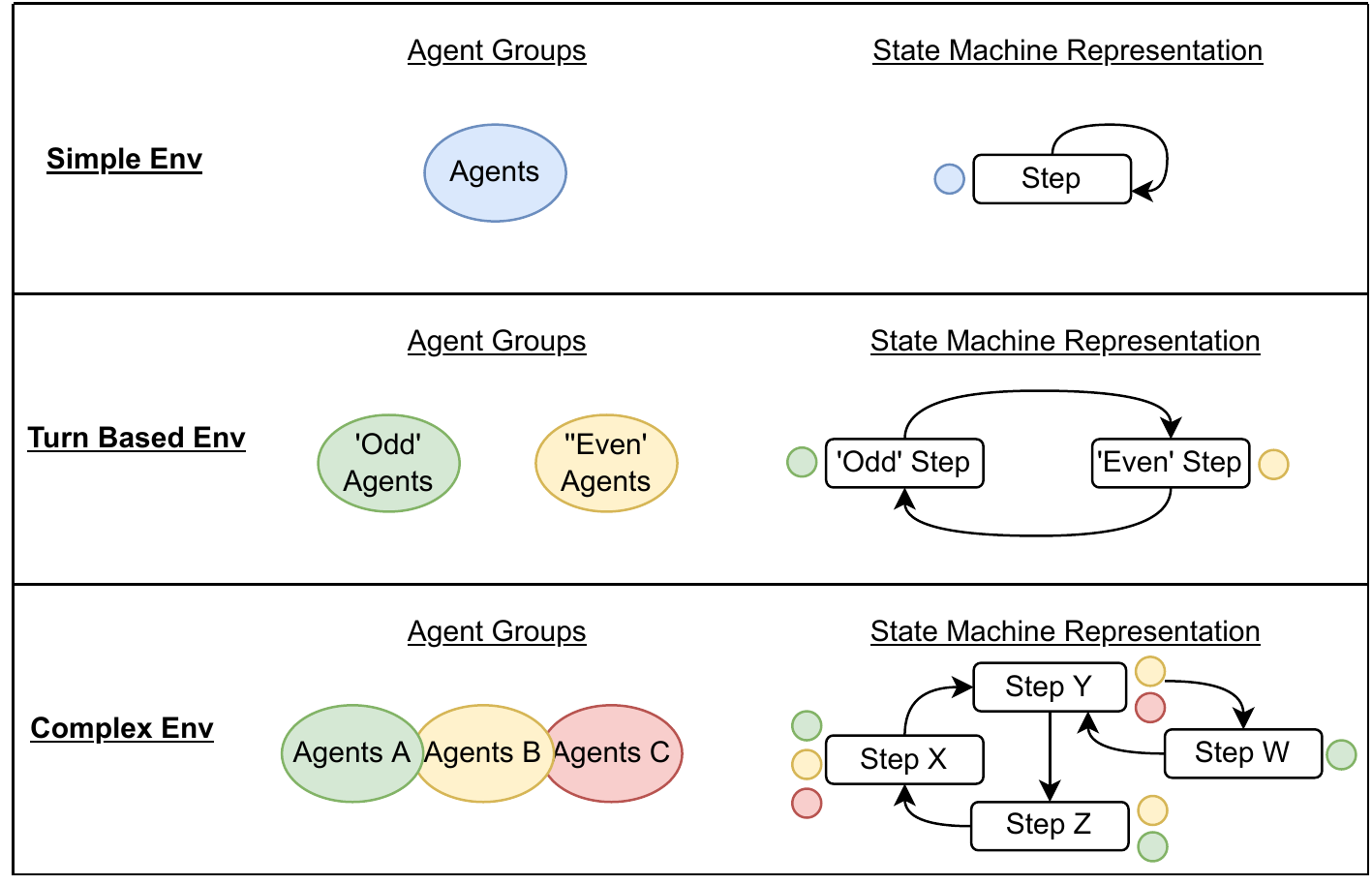}}
    \caption{State Machine Environment}
    \label{fig:env_finite_state_machine}
  \end{subfigure}
  \captionsetup{justification=centering}
  \caption{On the left, a breakdown of the different phases happening in an environment 'step'. On the right, some examples of environment designs, ordered by complexity. The State Machine Representation shows how the environment will orchestrate the simulation.}
  \Description{On the left, the sequence diagram of the different stages composing a "step". On the right, different examples of environment the \phantomName{} framework allows to design.}
  \label{fig:environment}
\end{minipage}
\end{figure*}

While the Types construct is powerful, it is still difficult to scale for a large number of agents. We argue that it is usually easier to consider families of agents sharing the same average behavior or "persona", as it is often named in the industry. Of course, each individual is unique and has its own characteristics but agents in the same persona tend to behave similarly on average. More concretely, a family of agents can be defined as a distribution over the Type parameter space - in \phantomName{} we refer to this distribution as a \textit{Supertype} as originally introduced in \citet{vadori2020calibration}.

At the beginning of each episode, a Type value is drawn from the Supertype for each agent, leading to variability in the agent configuration throughout the experiment (both training and simulation). With many episodes, the system simulated covers a variety of agent types and configurations, covering the set of behaviors of a given persona. 


The benefit of using Supertypes is two-fold. First, the formulation of the system is much more compact. Let us assume that each agent is defined by a single parameter $\gamma \in [0, 1]$. For $40$ agents in the system organised in $2$ personas, one where the parameter follows a uniform distribution between $0$ and $0.5$ $(\gamma \sim U(0.0, 0.5))$ and another one where $\gamma \sim U(0.5, 1.0)$. Without Supertypes, one would need to manually specify the $40$ $\gamma$ parameters for all the agents. On the other hand, with our approach we simply need to define $2$ distinct Supertypes generating $20$ agents each with the parameter drawn from the appropriate uniform distribution. That would account for: $2$ parameters for the uniform distribution plus the number of agents (i.e $3$ parameters) for each Supertype or only $6$ in total. Furthermore, the compactness of the system's formulation reduces the number of parameters to calibrate \citep{vadori2020calibration} making it easier to produce a system as close as possible to the real-world. Second, the built-in implementation of the Supertypes in our framework allows us to re-sample the agents' configuration after every episode, guaranteeing that for a number of episodes large enough, the full spectrum of behaviors will be exposed to the learning agents. This has the advantage of avoiding over-fitting to a particular configuration and making the policy able to cope with diverse conditions during inference.

\subsection{Modeling Complex Games}

We build upon the, now standard, Open-AI gym \citep{Brockman_Cheung_Pettersson_Schneider_Schulman_Tang_Zaremba_2016} paradigm where a learning agent interacts in discrete time with the rest of the system via the intermediary of a centralized \textit{environment}. The multi-agent setting adds a certain level of complexity to the environment component who now plays the role of orchestrator of the simulation. It is in charge of deciding when and in which order the agents get to act in the environment, when to send the messages between agents and when to update the agents' internal state. This complex logic is encapsulated under the `step` method of the environment interface. We present in Figure~\ref{fig:step_breakdown} the breakdown of a simulation step using \phantomName{}. First the actions coming from the agents' policy are processed to create both the \textit{Messages} that will need to be sent to the other agents and the \textit{Mutations} that will need to be performed to update the agent's internal state. The messages are dispatched through the network via a customizable \textit{Message Resolver}, deciding the ordering in which the messages will be sent. The messages are posted on a queue which is consumed until exhaustion allowing multiple messages back and forth between agents. Once all the messages have been processed, the mutations are executed to update the agent's internal state. Finally, the reward associated with the performed action is computed and the current observations are collected to decide on the next action to take.

With multiple agents at play, the complexity of the `step` method can rapidly increase and it becomes harder to design complex problems. To alleviate this, \phantomName{} provides a simple and modular way to implement complicated sequences of stages where only a subset of the agents act. It uses the \textit{Finite State Machine} formalism to define the order in which the agents are required to execute their actions in the environment. We show on Figure~\ref{fig:env_finite_state_machine}, different examples of environment going from a simple one-step environment to a more sophisticated one involving a complex sequence of stages. For instance, the "Turn Based Env" characterizes a Stackelberg game where the agents are categorized into two groups playing alternatively. This type of game can be used to evaluate how one group of agents react to the actions performed by the agents from the other group.

\subsection{Leveraging MARL for Scale}

\phantomName{} is setup to leverage advances in MARL and enable the use of RL-driven agent-based modeling at scale. 
The framework provides a direct integration with the distributed RL library RLLib \citep{pmlr-v80-liang18b}, providing scalable implementations of state of the art RL algorithms and allowing the study of large multi-agent systems. While \phantomName{} is natively integrated with RLLib, the framework was designed to be modular and unopinionated to allow the integration with other RL frameworks such as Stable Baseline 3 \citep{stable-baselines3} more suited for rapid prototyping.


The framework also offers a built-in implementation of the \textit{Shared Policy} learning technique presented in \citet{vadori2020calibration}, that can easily be configured via the framework's API. \phantomName{} automatically augments the observation space of an agent with its Type parameters for each episode, making it seamless to train policies that generalize across the range of values characterizing a family/supertype. It also allows the agents from the same family to share the same policy, considerably limiting the number of models to train.

There is a tremendous push towards learning the equilibria of complex general sum games with techniques that combine elements of game theory and RL. This is a rapidly advancing field and our hope is that with a RL-driven ABM framework like \phantomName, the ABM community can leverage these advances to realistically simulate complex domains. 

\section{Experiments}
In this section, we illustrate how the unique features offered by \phantomName{} can be used to simulate two distinct environments and help learn policies to act optimally in a multi-agent setting.

\subsection{Digital Ads Market Environment} \label{sec:digital_ads_env}

We use \phantomName{} to design an environment aiming to simulate the digital ads market \footnote{See \cite{https://doi.org/10.48550/arxiv.1204.0535} for more information about the Digital Ad Market.} where \textit{Advertisers} (e.g an Airline company) compete in an auction to win the right to display their ads to some \textit{Users} visiting a website (the \textit{Publisher}). Each User has different interests such as travel, sport, tech etc., making them more likely to click on the ads inline with their preferences (e.g someone highly interested in travel has a higher probability of clicking on an Airline discount ad than someone preferring to read tech news). Every time the User accesses a website an "impression request" is sent to the Advertisers with information about the User. At this stage, the Advertisers needs to decide how much they want to bid to display their ad to this User. The bids from all the Advertisers are collected by an \textit{AdExchange} in charge of running the auction and deciding which Advertiser has won the right to show their ad. Once the User is presented with the ad, she clicks with a certain probability on it, giving the opportunity to the Advertiser to make a sale. This problem can be seen as a constrained optimization problem for the Advertisers as they need to decide how much they want to bid on each impression while being subject to a fixed budget to bid with. The Advertisers want to bid high enough to win the auction on Users with a higher probability of clicking on their ad (without knowing such probability), but they also want to be mindful on how much they bid to maximize the number of impressions they make and consequently increase the number of clicks they get.

In our set of experiments, we define 3 families of Advertisers, each with a different theme (Travel, Tech and Sport). Each family is composed of 2 agents. There is one central AdExchange and one Publisher simulating a website that can be accessed by 2 different Users. One User has a higher probability to click on Tech and Travel ads, while the other is more interested in Sport (Table~\ref{tab:digita_ads_user_probas}). The Travel and Tech Advertisers will therefore prefer to compete for User 1, while the Sport Advertiser will favour User 2. The budget allocated to each agent of the same family of Advertisers is sampled from a Uniform distribution. The Tech Advertisers' budget can range from $10$ to $20$, the Sport Advertisers from $7$ to $17$ and the Travel Advertisers from $5$ to $15$. We use \phantomName{} and the Shared Policy feature with the PPO implementation of RLLib \citep{pmlr-v80-liang18b} to train one policy for each of the Advertiser family. The full RL description of the agent is available in the supplementary material.

\begin{table}[ht]
  \captionsetup{justification=centering}
  \caption{Probability of each User to click on an Ad of a given theme.}
  \label{tab:digita_ads_user_probas}
  \begin{tabular}{rccc}\toprule
     & \textit{Travel} & \textit{Tech} & \textit{Sport} \\ \midrule
    User 1 & 1.0 & 0.8 & 0.0 \\
    User 2 & 0.0 & 0.2 & 1.0 \\ \bottomrule
  \end{tabular}
\end{table}

\subsubsection{\phantomName{} Features}
\hfill

\noindent\textbf{Messages and Views:} The \textit{Message} passing feature of \phantomName{} allows back and forth communication between the agents within the same timestep. The Advertiser is thus able to directly receive the reward associated with its action: if its bid was not high enough, the AdExchange will indicate the Advertiser that it did not win the auction; if the Advertiser placed a bid on a user with a low probability of clicking on the ad, the Publisher will inform the Advertiser of the User's action giving the Advertiser the opportunity to adjust its bidding strategy. The \textit{Views} are used to provide additional information to the Advertisers such as age or zipcode area, in order to help them make a better decision. The Views being customizable at the agent level, one can think of scenarios where additional information about the Users are available for a certain fees that Advertiser may or may not decide to pay, adding more heterogeneity in the system.

\noindent\textbf{Types and Supertypes:} To cover a range of possible bidding budgets allocated to Advertisers without having to re-train a policy for each possible value, we use the Supertype feature of \phantomName{} and define a distribution to sample from to set the budget of an Advertiser. During our analysis we will be able to evaluate various scenarios with different allocated budgets.

\noindent\textbf{Finite State Machine Environment:} To design the sequence of action in which agent can act in the environment, we use the implementation of the Finite State Machine environment provided by \phantomName{}. One simulation step is divided into two sub-steps. First, the Publisher acts by submitting an impression request to the AdExchange, which will forward it to the connected Advertisers. Then, the Advertisers acts by sending their bid to the AdExchange in charge of performing the auction; the winner Advertiser sends its ad to the Publisher which will reply with the results of the impression. This complex sequence of actions is easily implemented thanks to \phantomName{}.

\subsubsection{Emergent Behaviors}
\hfill

\begin{figure}[h]
    \captionsetup{justification=centering}
    \centering
    \centerline{\includegraphics[scale=0.43]{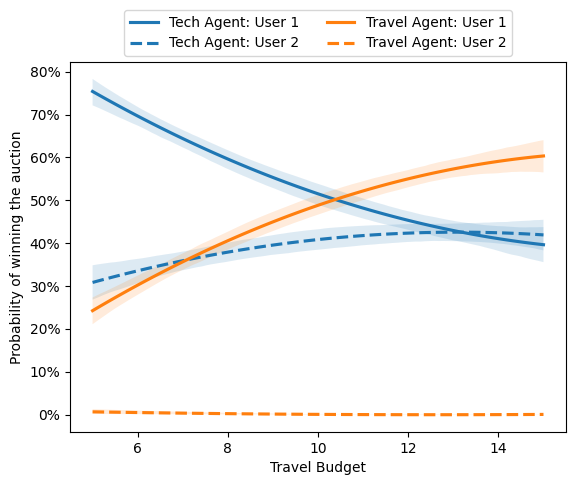}}
    \caption{Probability of winning the auction as a function of the Travel agent budget, for the competing Tech and Travel agent.}
    \label{fig:proba_winning}
\end{figure}

We analyze in Figure~\ref{fig:proba_winning} how a Tech Advertiser with a fixed budget of $10$ adapts its behavior when the budget allocated to its main competitor changes. We plot the probability of the Travel and Tech Advertisers to win the auction for both of the Users as a function of the Travel Advertiser's budget.

First, we see that the Travel agent has correctly learnt that bidding on User 2 was a waste of money, as its probability of clicking on a Travel Ad is null. The second observation, is that with a higher budget the Travel Advertiser is able to win the auctions involving User 1 with a higher probability. Indeed, with more budget, the Advertiser can afford to place higher bids and therefore win the auction more often. The most interesting insights, is the evolution of the Tech Advertiser's behavior. As the budget of the Travel Advertiser increases, the Tech Advertiser wins the auction with a lower probability. To compensate for the loss in the auctions for User 1, the Tech Advertiser shifts its strategy to bid more on the other User increasing its probability of winning the auctions for User 2.

\begin{figure}[h]
    \captionsetup{justification=centering}
    \centering
    \centerline{\includegraphics[scale=0.45]{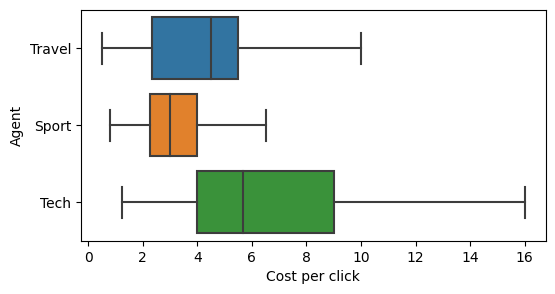}}
    \caption{Distribution of the cost per click for the 3 families of Advertisers.}
    \label{fig:cost_per_click}
\end{figure}

A typical measure used to evaluate the performance of the Advertiser's bidding strategy is called the "cost per click" and corresponds to the average money spent to get a User to click on an Ad. We present in Figure~\ref{fig:cost_per_click} the distribution of the cost per click measure for the 3 families of Advertisers.

The Travel and Tech Advertisers are mainly competing for User 1, as it has the highest probability of clicking on their ads. The Sport Advertiser however, is the only Advertiser focusing on User 2, for this reason its "cost per click" is lower than the two others. Despite having a lower budget the Travel Advertiser has on average a lower cost per click due to the fact that User 1 has a probability of $1.0$ of clicking on a Travel Ad but only $0.8$ of clicking on a Tech Ad.

\subsection{Scalability Experiments}
To evaluate the scalability of \phantomName, we use the Digital Ads Market environment as a testbed to scale the number of agents present in the system and observe the performance. We make use of the Shared Policy feature to learn a reduced number of policies able to generalize across a range of different behaviors allowing the system to scale.

We use the configuration presented in Section~\ref{sec:digital_ads_env} where $3$ families of Advertisers, each with their own theme, compete to show their ads to $2$ different users. The budget allocated for each agent of a given family is sampled from the same distribution allowing heterogeneity in the behaviors adopted by the agents while remaining similar on average. For these experiments, we increase the number of agents in each family  agents present in the system. The experiments are run on a \textit{c5.12xlarge} EC2 instance with 48 vCPUs and 96GB of RAM using RLLib \citep{pmlr-v80-liang18b} with $40$ workers.

\begin{figure}[h]
    \captionsetup{justification=centering}
    \begin{subfigure}[b]{0.27\textwidth}
        \centering
        \includegraphics[width=\textwidth]{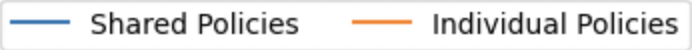}
    \end{subfigure}
    \hfill
    \begin{subfigure}[b]{0.22\textwidth}
        \centering
        \includegraphics[width=\textwidth]{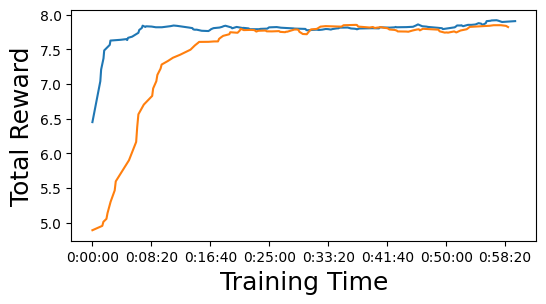}
        \captionsetup{justification=centering}
        \caption{Time to convergence} 
        \label{fig:time_to_convergence}
    \end{subfigure}
    \begin{subfigure}[b]{0.22\textwidth}
        \centering
        \includegraphics[width=\textwidth]{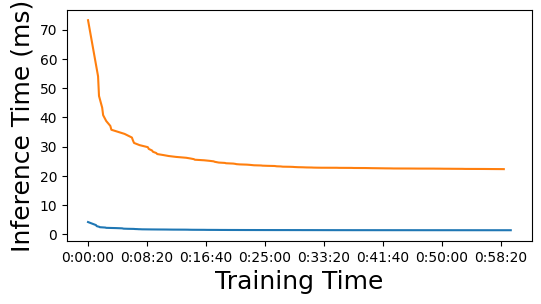}
        \captionsetup{justification=centering}
        \caption{Inference Time} 
        \label{fig:inference_time}
    \end{subfigure}
    \begin{subfigure}[b]{0.22\textwidth}
        \centering
        \includegraphics[width=\textwidth]{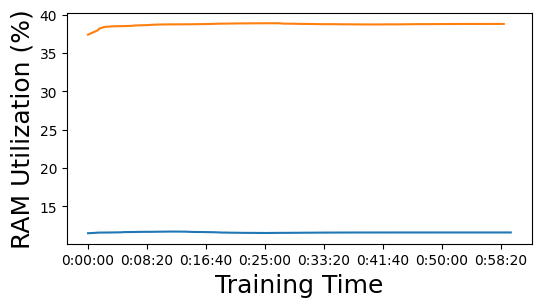}
        \captionsetup{justification=centering}
        \caption{RAM Utilization} 
        \label{fig:ram_util}
    \end{subfigure}
    \begin{subfigure}[b]{0.22\textwidth}
        \centering
        \includegraphics[width=\textwidth]{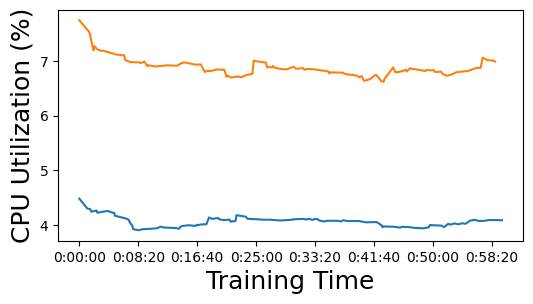}
        \captionsetup{justification=centering}
        \caption{CPU Utilization} 
        \label{fig:cpu_util}
    \end{subfigure}
    \caption{Scalability metrics with and without using the Shared Policy learning feature offered by \phantomName{}.}
    \label{fig:scalability}
\end{figure}

In Figure~\ref{fig:scalability} we evaluate the impact of using the Shared Policy feature of \phantomName{} when we have $60$ learning agents in the system ($20$ for each Advertisers family). We see that it improves the overall performance of the simulation. The trained policies reach convergence significantly faster when the Shared Policy feature is used ($\sim8$min vs $\sim22$min), but more importantly the amount of computing resources required to train the agents is much less allowing the simulation of bigger systems. With the \phantomName{} Shared Policy feature, training the $60$ agents requires almost $3.5$ times less RAM than without.

 We present in Table~\ref{tab:training_time} the performance metrics for an increasing number of learning agents in the system.\footnote{More details are presented in the supplementary materials.}

\begin{table}[ht]
  \captionsetup{justification=centering}
  \caption{Performance metrics for an increasing number of learning agents.}
  \label{tab:training_time}
  \begin{tabular}{rcccc}\toprule
    & \shortstack{6\\agents} & \shortstack{60\\agents} & \shortstack{120\\agents} & \shortstack{1200\\agents} \\ \midrule
    Time to converge & 6min & 8min & 15min & 4h10min \\ \bottomrule
  \end{tabular}
\end{table}

\subsection{Supply Chain Environment} 
\label{sec:supply_chain_env}
Supply chain management literature reflects that significant efforts have been put into deriving stock optimization strategies to achieve empirically provable optimality, but at the cost of simplifying the problem formulation due to dearth of open-source tools available to achieve authentic simulation and mechanism design of such environments \citep{humair2013incorporating, de2018typology}. 

Our supply chain environment settings consist multiple shops learning an effective inventory restock strategy, based on dynamically varying conditions such as number of customer orders that they receive and inventory storage costs etc. We simulate 3 family of agents; Customer, Shop and Factory. The shop is the only learning agent, while customer and factory are the non-learning agents. At each time-step, the customers place variable quantity of stock order (based on discrete uniform distribution) to different shops. Figure ~\ref{fig:messages-views} describes the mechanics of this environment. Depending on the availability of stock, the shops fulfil the orders. We use an RL-based method (PPO \citep{schulman2017proximal}) to learn a shared policy. 
The shop agent's observation space consists of values from the previous time-step: customer orders received, sales made, leftover (unsold) stock and cost of carry (characterizing the "type" of the agent). The agent decides on the restock quantity to order at the factory. Our reward design includes the revenue generated minus the cost incurred. More details are available in the supplementary materials.



\subsubsection{\phantomName{} Features}
\hfill
 
\noindent\textbf{Messages and Views:}  \phantomName{} allows customised and secure connectivity/communication between different agents using the message passing mechanism, so that only the agents that are supposed to be connected connect and not otherwise. In our supply chain environment, the customers and shops are connected to respectively place orders and get them fulfilled from shops. Additionally, there exists connectivity between the shop and factory agents for stock replenishment. \phantomName{}'s View feature allows the shops to broadcast their varied selling price to the customers so that the customers can choose which shop to buy stock from. 

\begin{figure}[ht]
    \captionsetup{justification=centering}
    \centering
    \centerline{\includegraphics[scale=0.43]{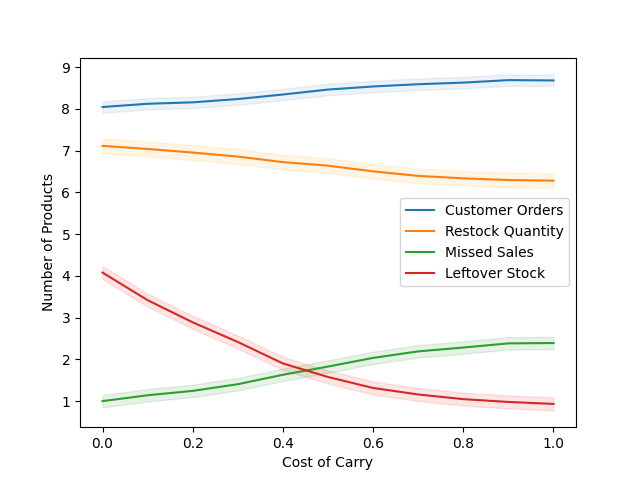}}
    \caption{The impact of increasing cost of carry on relevant metrics.}
    \label{fig:cost_of_carry}
\end{figure}

\noindent\textbf{Types and Supertypes:} \phantomName{}'s supertype feature is used to enable the shop agent to generalise to a range cost of carry values and learn a robust restock policy which can effectively accommodate to varied conditions resulting from variation in cost of carry values. For instance, at a high cost of carry, we expect the shop agent to restock conservatively to have minimum leftover (unsold) stock as there is a high cost associated with storing the leftover stock (Figure~\ref{fig:cost_of_carry} reflects this behavior).  

\noindent\textbf{Finite State Machine Environment:} Using this feature, we segregate the sale and the restock step. In the first time-step, the restocking task is accomplished and in the subsequent time-step, the task of receiving customer orders and selling stock is performed. This pattern is repeated thereafter until the shop agent achieves convergence. This way of sequentially handling events contributes to simplicity and tractability that is often needed to simulate complex environments involving multiple acting agents. 


\begin{figure}[ht]
    \captionsetup{justification=centering}
    \centering
    \centerline{\includegraphics[scale=0.43]{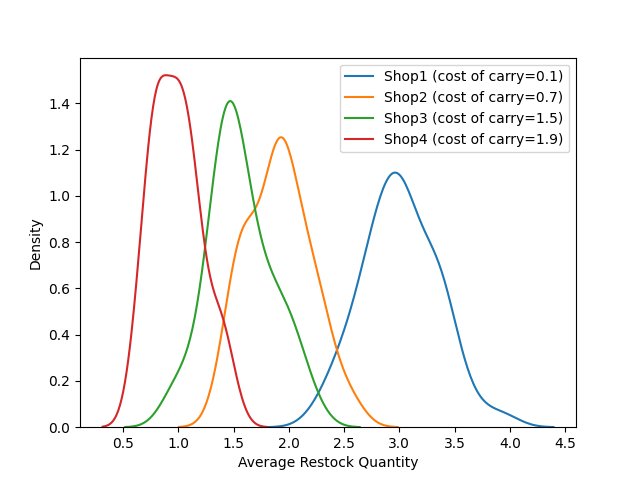}}
    \caption{Distribution of average restock quantity for 4 shops, with shop 1 having the lowest cost of carry and shop 4 having the highest cost of carry.}
    \label{fig:cost_of_carry1}
\end{figure}

\subsubsection{Emergent Behaviors}
\hfill

\begin{table*}[h!]
    \captionsetup{justification=centering}
    \caption{Comparison of different ABM and MARL frameworks}
    \label{tab:comparison_frameworks}
    \begin{tabular}{ccccc}\toprule
    & \textit{NetLogo} & \textit{WarpDrive} & \textit{MAVA} & \textbf{\phantomName}\\\midrule
    Discrete-Event Simulation & \ding{51} & \ding{51} & \ding{51} & \ding{51} \\
    Support Learning Agents & \ding{55} & \ding{51} & \ding{51} & \ding{51} \\
    Stochastic Network & \ding{55} & \ding{55} & \ding{55} & \ding{51} \\
    Enforced Partial Observability & \ding{55} & \ding{55} & \ding{55} & \ding{51} \\
    Complex turn-taking environments & \ding{55} & \ding{55} & \ding{55} & \ding{51} \\
    Optimized RL Algorithms & \ding{55} & \ding{51} & \ding{51} & \ding{55}* \\\bottomrule
    \end{tabular}\\
    \small{* Available when combined with RLLib \citep{pmlr-v80-liang18b}.}
    \label{table:ABMcomparison}
\end{table*}

In this section, we demonstrate the interesting behavior that emerges through the various dynamic interactions that take place in our supply chain environment. An illustration of the impact of increasing cost of carry on various metrics is shown in Figure~\ref{fig:cost_of_carry} in the evaluation/testing phase of our experiment (trained policy is used). As the cost of carry increases, the shop agent gets conservative in restocking in order to have minimum leftover stock as there is a high cost associated with storing the leftover stock, but at the cost of some missed sales. 

Specifying a separate supertype (i.e. a distribution to sample from) for all the learning agents can be resource-intensive. To avoid overhead, only one agent can be trained with supertype and we expect the other agents to generalise. Figure ~\ref{fig:cost_of_carry1} depicts this behavior. We train one shop with cost of carry supertype range $0-2$. Then, during the evaluation, we increase the number of shops to 4 and we fix different cost of carry for different shops (shop1$=0.1$, shop2$=0.7$, shop$3=1.5$ and shop$4=1.9$). As shown in Figure ~\ref{fig:cost_of_carry1}, we can see a shift (on the left) in average restock quantity as the cost of carry increases (i.e. the shop with the lowest cost of carry restocks the most, while the shop with the highest cost of carry restocks the least).





\section{Related Work}

Despite having been around since the 70's \citep{schelling_DynamicModels_71}, the notion of Agent Based modeling is an area that really started to grow in the 90's This sudden expansion can in part be attributed to the development of multi-agents frameworks such as \textit{SWARM} \citep{swarm}, \textit{NetLogo} \citep{tisue2004netlogo} and others, making ABM more accessible to practitioners, reducing the barriers to entry in the field. Since then we have seen numerous applications of ABM in a variety of fields: flow simulation \citep{Helbing2000SimulatingDF}, markets simulation \citep{Palmer1994ArtificialEL}, organizational simulations \citep{Prietula1998SimulatingOC}. 

These frameworks, built quite some time ago, have helped the research community study complex systems but are not natively geared towards leveraging MARL. \textit{NetLogo} \citep{tisue2004netlogo} is built on top of Java but provides its own programming language. The framework was developed with performance in mind to be able to support a large number of agents exchanging many messages. Although, Java is a proven language to deal with low-latency systems, its verbosity and its complexity prevent a broad adoption from the AI research community. On the other hand, the Python language, owing to its simplicity and the ease to build and import new libraries, has established itself as the go-to language for Machine Learning and other AI sub-fields. Among ABM frameworks, \textit{ABIDES} \citep{abides} is a discrete event simulator with a Python interface that has been successfully used to model limit-order books in finance, but it relies on hand-coded agent policies. 

In most recent years, we have seen an increase in the development of RL-frameworks designed for fast code iteration and rapid experimentation. In \citeyear{TFAgents}, \textit{TF-Agents} was created as an additional module to the \textit{TensorFlow} framework to "make implementing, deploying, and testing new Bandits and RL algorithms easier" \citep{TFAgents}. The concept of agent is introduced as a core element of the module. However, although it supports a multi-agent setting, the framework was not designed with this in mind making the implementation of multi-agents system more convoluted.


MARL frameworks such as \textit{WarpDrive} \citep{warpDrive} and \textit{MAVA} \citep{MAVA_2021} are designed to enable easier and more efficient implementation of MARL algorithms. The former innovates by focusing on performance with the use of GPU and their parallelization power. The multi-agent RL framework runs the simulation on a single GPU avoiding transferring data between the rollout workers and the policy trainer. \textit{MAVA} also proposes a new distributed framework for multi-agent RL. It leverages many of Deepmind's open source components such as Acme the distributed single agent RL framework \citep{hoffman2020acme}, Reverb for data management \citep{reverb} and Launchpad for distributed processing orchestration \citep{yang2021launchpad}. Like \phantomName, \textit{MAVA} offers the options to specify network configuration to model the agents communication, however unlike in \phantomName{} the network configuration remains fairly basic and stays static throughout the simulation and therefore does not allow the study of systems with stochastic connectivity. 

We present in Table~\ref{table:ABMcomparison}, a brief comparison of \phantomName{} with other ABM and MARL frameworks. More broadly, in contrast to the RL and MARL frameworks, \phantomName's goal is to provide the tools to develop the multi-agent \textit{environment} for a problem or domain (as an ABM), rather than MARL algorithm itself. Indeed, our hope is that \phantomName{} enables the ABM community to easily leverage the power of new MARL frameworks to learn agent behaviors at scale.

\section{Conclusion}
 In this paper, we introduced a new framework, \phantomName{} that leverages the power of MARL to automatically learn agent behaviors or policies, and the equilibria of complex general-sum games. To enable this, the framework provides tools to specify the ABM in MARL-compatible terms - including features to encode dynamic partial observability, agent utility / reward functions, heterogeneity in agent preferences or types, and constraints on the order in which agents can act. We presented the rationale behind the implementation of the main features built to ease the development of complex multi-agent systems. We also provided some examples where the framework has helped analyze a system such as the financial market, well-known for its complexity.  Our hope is that \phantomName{} enables the ABM community to conveniently leverage the power of MARL algorithms to learn complex and realistic agent behaviors at scale. 

\balance
\newpage
\renewcommand\acksname{Disclaimer}
\begin{acks}
This paper was prepared for informational purposes by the Artificial Intelligence Research group of JPMorgan Chase \& Co and its affiliates (“J.P. Morgan”), and is not a product of the Research Department of J.P. Morgan. J.P. Morgan makes no representation and warranty whatsoever and disclaims all liability, for the completeness, accuracy or reliability of the information contained herein. This document is not intended as investment research or investment advice, or a recommendation, offer or solicitation for the purchase or sale of any security, financial instrument, financial product or service, or to be used in any way for evaluating the merits of participating in any transaction, and shall not constitute a solicitation under any jurisdiction or to any person, if such solicitation under such jurisdiction or to such person would be unlawful.
© 2022 JPMorgan Chase \& Co. All rights reserved.
\end{acks}

\bibliographystyle{ACM-Reference-Format}
\bibliography{references}


\end{document}